\documentclass[fleqn,12pt]{article}
\usepackage{espcrc1}

\usepackage{amssymb,amsmath}
\usepackage{lscape,graphicx}
\usepackage{algorithmic}
\usepackage{algorithm}
\usepackage{rotating}
\usepackage{subfigure}
\usepackage{times}

\title{Automatic Construction of Lightweight Domain Ontologies for Chemical Engineering Risk Management}

\author{
W. Wong, W. Liu, $^1$S. Liaw, $^2$N. Balliu, $^2$H. Wu, $^2$M.O. Tade\\
University of Western Australia, School of Computer Science and Software Engineering, 35 Stirling Highway, Crawley, WA 6009, Australia; tel. +61-8-64883095, fax. +61-8-64881089 e-mail: \{wilson,wei\}@csse.uwa.edu.au,$^1$saujoe@gmail.com\\
$^2$Curtin University of Technology, Department of Chemical Engineering, GPO Box U1987, Perth, WA 6001, Australia; tel. +61-8-9266 7581, fax. +61-8-92662681, email: \{N.Balliu,H.Wu,M.O.Tade\}@curtin.edu.au
}

\begin{document}

\maketitle

\begin{abstract}
The need for domain ontologies in mission critical applications such as risk management and hazard identification is becoming more and more pressing. Most research on ontology learning conducted in the academia remains unrealistic for real-world applications. One of the main problems is the dependence on non-incremental, rare knowledge and textual resources, and manually-crafted patterns and rules. This paper reports work in progress aiming to address such undesirable dependencies during ontology construction. Initial experiments using a working prototype of the system revealed promising potentials in automatically constructing high-quality domain ontologies using real-world texts.
\end{abstract}

\section{Introduction}
Hazard identification is a crucial aspect of risk management. The identification of hazards is the prerequisite step to the analysis and treatment of risks. As such, clear definitions on the type of risks and the processes involved for hazard avoidance and treatment are necessary. Unambiguous definition enables effective communication, which is crucial in passing on experiences and expertise to trainees and students dealing with dangerous chemicals and products. However, very often such knowledge is embedded in the domain experts' mind, or scattered in various format, e.g. operation notes, online resources, scientific publications or technical reports. An integrated knowledge structure known as an ontology is therefore becoming necessary for describing the concepts and processes to ease the process of information sharing and reuse. Some possible applications of domain ontologies include conceptual document retrieval and decision support system. The importance of ontologies to knowledge-based applications has prompted an increase in efforts to construct and maintain such knowledge structures. Generally, there are two ways of constructing ontologies, namely, manual crafting and automatic discovery. 

Manual construction and maintenance of ontology is often critised for being labour intensive, biased and static. Such manual process typically requires multiple domain experts to identify the key concepts and processes, and then collaborate with knowledge engineers for effective digital representation. The neutrality and representativeness of manually-crafted ontologies is also disputable when the domain experts are unable to reach consensus during the knowledge engineering process. New changes to the domain are often ignored and cannot be incorporated into the ontology in a timely manner. To address the problems related to manual ontology construction and maintenance, several systems have been developed in the past for automatically constructing domain ontologies. However, these existing systems are mainly based on manually-crafted patterns and rules, and non-incremental, static textual resources. Such dependencies impose great restrictions on the systems' applicability to a wider range of domains. Moreover, automatic ontology construction is a relatively new research area. Many techniques used in existing systems were borrowed from related fields in Computer Science such as Information Extraction, Information Retrieval and Text Mining. Direct applications of such techniques are inadequate in addressing the various peculiarities of converting real-world natural language texts to domain ontologies. Ideally, systems should focus on automatically generating high-quality lightweight domain ontologies with facilities for manual refinements by domain experts.

In this paper, we introduce a system for automatic ontology construction to promote the establishment of a standard vocabulary in the field of risk management. To address the above issues, our ontology construction system utilises dedicated techniques for constructing lightweight ontologies, relying only on dynamic textual resources on the Web. The system extracts key concepts and relations automatically from electronic domain texts to construct domain ontologies. The domain texts may be technical reports, operation notes, electronic books and web pages covering risk management and hazard identification in chemical engineering processes. The lightweight ontology can then be maintained or edited by domain experts using various ontology editing software. The paper is organised as follows. We provide a brief review of related work in Section \ref{relatedwork}. We proceed to elaborate on the system architecture and its strengths in Section \ref{systemarchi}. In Section \ref{experiment}, we demonstrate the process of ontology construction by conducting an experiment with a working prototype of our system using domain texts constructed from ScienceDirect and a textbook. In Section \ref{conclude}, we conclude with an outlook to future work.

\section{Related Work}\label{relatedwork}

In this section, we briefly review some manual efforts and automated systems for constructing ontologies. Prior to the rise in popularity of automatic ontology construction, domain experts engaged in collaborative efforts to create ontologies. One of such pioneering projects is the \emph{Gene Ontology (GO)} \cite{ashburner_et_al_2000}. Even to date, the impractical constraints imposed by existing automatic ontology construction systems and their far from satisfactory results have prompted experts to continue working manually to construct high-quality domain ontologies. The \emph{Plant Ontology Consortium (POC)} \cite{jaiswal_et_al_2005} is one of the more recent handcrafted ontologies which integrates a wide range of vocabularies used to describe the anatomy, morphology and growth stages of several plants. The \emph{European Bioinformatics Institute (EBI)} initiated a collective effort to construct the \emph{Chemical Entities of Biological Interest (ChEBI)} ontology \cite{degtyarenko_et_al_2008}. \emph{ChEBI} is an ontology which focuses on molecular entities used to intervene in the processes of living organisms. Many of these manual efforts are possible through the widely available ontology development tools such as \emph{OntoLingua} \cite{farquhar_et_al_1995} and $Prot\acute{e}g\acute{e}$ \cite{musen_et_al_1993}. In the related domain of risk management, a recent work by \cite{gilmour_2004} produced a domain ontology through the manual identification and organisation of key concepts and processes for the domain of hazard identification using $Prot\acute{e}g\acute{e}$. 

Besides manual efforts, several ontology construction systems have also been developed in recent years which aim at generating domain ontologies. For example, \emph{OntoLearn} \cite{velardi_et_al_2005} employs standard natural language processing (NLP) tools and corpus analysis to extract and recognise domain terms. \emph{Lexico-syntactic patterns} \cite{hearst_1998} and \emph{WordNet} \cite{miller_et_al_1990} are utilised to extract semantic relations between the terms. Similarly, the \emph{Text-to-Onto} system \cite{cimiano_staab_2005} makes use of non-incremental resources such as \emph{WordNet}, and manually-crafted lexico-syntactic patterns to construct ontologies. In order to identify more complex relations, \emph{Text-to-Onto} employs association rule learning. More recent work from \cite{jiang_et_al_2007} extract terms and semantic relations through dependency structure analysis. The terms are mapped onto \emph{WordNet} to obtain bags of senses. These senses are then clustered using cosine similarity. Semantic relations that consist of similar terms can be generalised using association rule mining algorithms for deducing statistically significant patterns. \cite{rozenfeld_feldman_2007} conducted a study on clustering and the associated tasks of feature extraction and selection, and similarity measurement for constructing ontologies. Contexts, appearing as sentences in which the terms occur, are used as features in their study. \cite{ramakrishnan_et_al_2006} utilise dependency structure analysis to extract terms and relationships with the help of a controlled vocabulary called the \emph{Medical Subject Headings (MeSH)} and domain knowledge in the form of the \emph{Unified Medical Language System (UMLS)}. 

The reliance on non-incremental lexical resources (e.g. \emph{MeSH} and \emph{WordNet}), and the use of non-dedicated techniques makes such ontology construction systems inapplicable to real-world applications and not portable to other domains.

\section{System Architecture}\label{systemarchi}

\begin{figure}[t]
    \begin{center}
\includegraphics[width=4.3in]{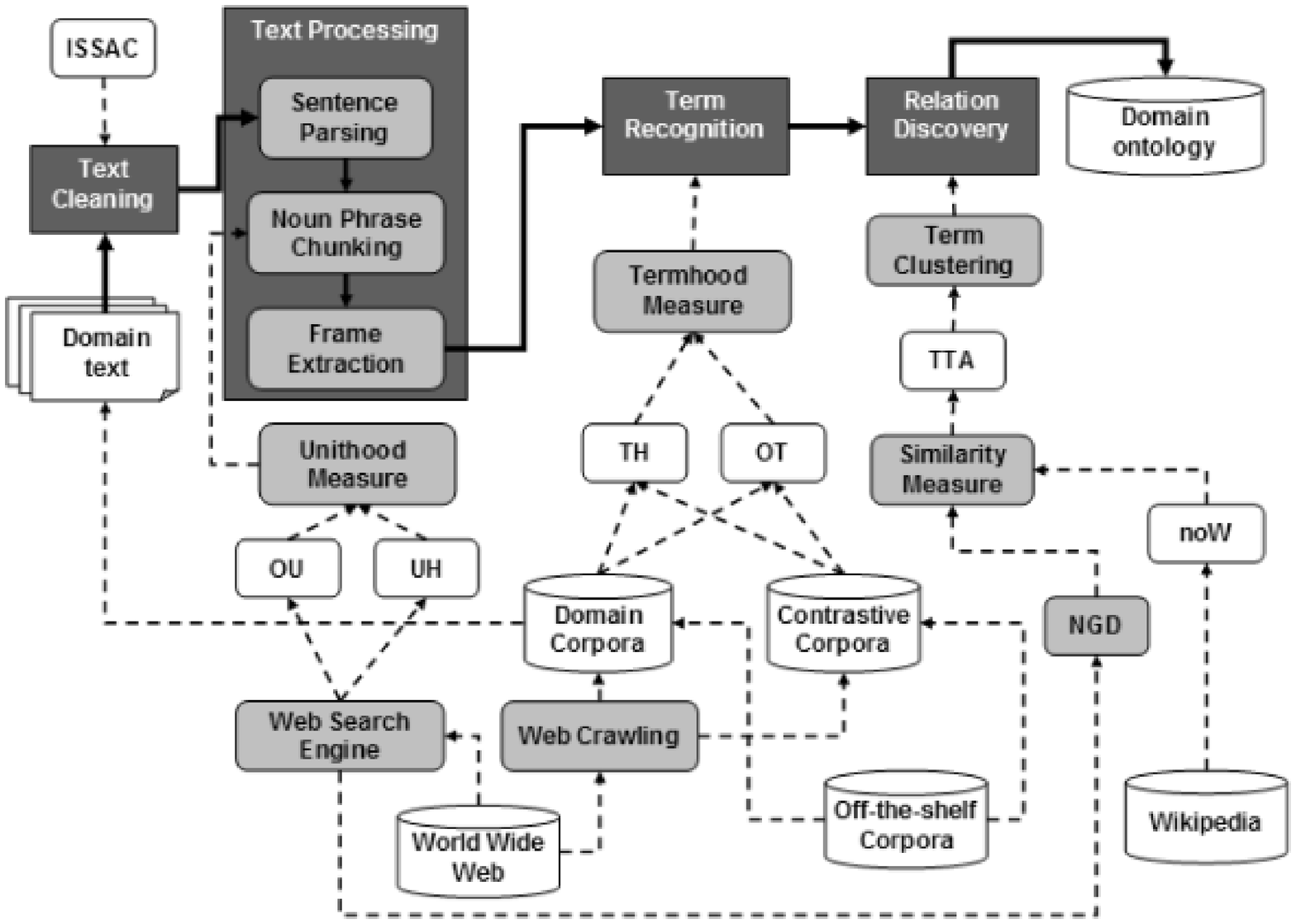}
        \caption{Ontology construction system architecture. The main phases, namely, \emph{text cleaning}, \emph{text processing}, \emph{term recognition} and \emph{relation discovery} in the system are represented using dark rectangles. The grey rounded rectangles represent existing techniques and functionalities required by the system while the white rounded rectangles depict novel techniques developed in this research.}\label{architecture}
    \end{center}
\end{figure}

\begin{figure}[t]
    \begin{center}
\includegraphics[width=5in]{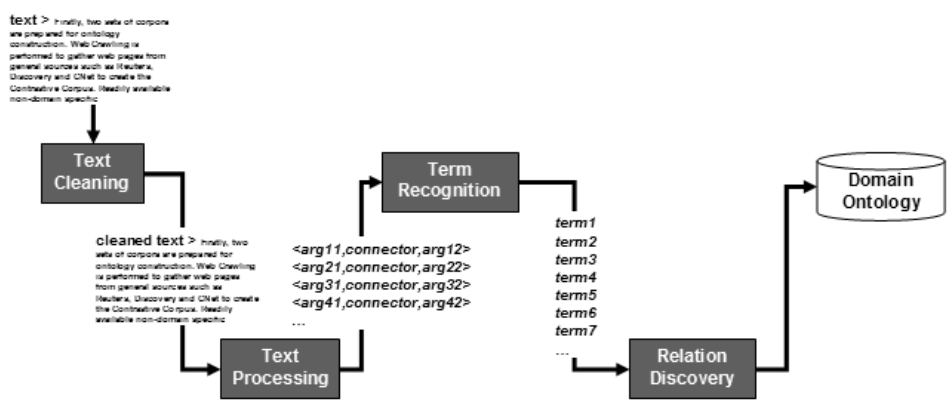}
        \caption{The intermediate output produced by the system at each phase. The text processing phase produces ternary frames in the form of $<arg1,connector,arg2>$ from natural language texts. The term recognition phase then identifies domain-relevant terms using the frames. Finally, the relation discovery phase identifies associations between the domain terms for constructing a lightweight domain ontology.}\label{outputflow}
    \end{center}
\end{figure}

In this section, we present the architecture of our ontology construction system which is designed to overcome the use of non-dedicated techniques, and the reliance on non-incremental resources and manually-crafted patterns and rules. The system is comprised of four main phases as shown in Figure \ref{architecture} (using shaded rectangles). These phases are \emph{text cleaning}, \emph{text processing}, \emph{term recognition} and \emph{relation discovery}. Text cleaning removes noises such as spelling errors, abbreviations and improper casings from texts. Text processing then extracts coherent three-part structures from the texts using linguistic information. Term recognition uses the extracted structures to produce a list of term candidates which are then shortlisted based on their relevance to the domain of interest. During the last phase of relation discovery, the semantic relations between terms are discovered to construct a domain ontology. The flow of intermediate output produced after each phase of processing is shown in Figure \ref{outputflow}. As we will show in Section \ref{experiment}, errors introduced at each phase have effects on the performance of subsequent phases. The specific functionalities required in each phase are shown using rounded rectangles in Figure \ref{architecture}. Several techniques developed as part of this research for ontology construction are shown using white rounded rectangles while existing techniques and resources required by the system are depicted as shaded rounded rectangles. Unlike conventional ontology construction systems, our techniques were specifically developed to handle the peculiarities of the input at each phase. We will elaborate more on the innovative aspects of our techniques as we progress along. Besides the techniques, the preparation of text corpora is equally important. The system requires two sets of text corpora, namely, a \emph{contrastive corpus} and a \emph{domain corpus}, which are depicted as cylinders in Figure \ref{architecture}. The contrastive corpus is populated with general and non-domain specific electronic collections of texts, and web pages obtained from general news sites through \emph{web crawling}. The domain corpus is built through guided web crawling which harvests scientific publications on the Web using key phrases provided by domain experts. Electronic versions of domain documents such as textbooks can also be added to the domain corpus. 

Ontology construction begins by performing an optional text cleaning phase using our technique known as \emph{Integrated Scoring for Spelling error correction, Abbreviation Expansion and Case restoration (ISSAC)} \cite{wong_et_al_2006b,wong_et_al_2007a}. \emph{ISSAC} is built upon the famous spell checker Aspell \cite{atkinson_2006} for simultaneously providing solution to spelling errors, abbreviations and improper casing. The technique has demonstrated high accuracy \cite{wong_et_al_2006b,wong_et_al_2007a} in correcting these three types of noises. \emph{ISSAC} combines weights based on various sources such as online abbreviation dictionary\footnote{http://www.acronymfinder.com/}, string correction algorithm \cite{wagner_fischer_1974}, contextual information, and co-occurrence analysis to determine the best replacement for each error word. Texts from edited or reputable sources such as academic journals can bypass this text cleaning phase. Many linguistic analysis tools and the systems that rely on them assume that the input texts are free from spelling errors, abbreviations and improper casings. However, the presence of such noises is inevitable in real-world texts, especially those from online sources. The inclusion of a text cleaning phase in our ontology construction system ensures robustness and better performance when dealing with online texts.

\begin{figure}[thp]
    \begin{center}
\includegraphics[width=3.2in]{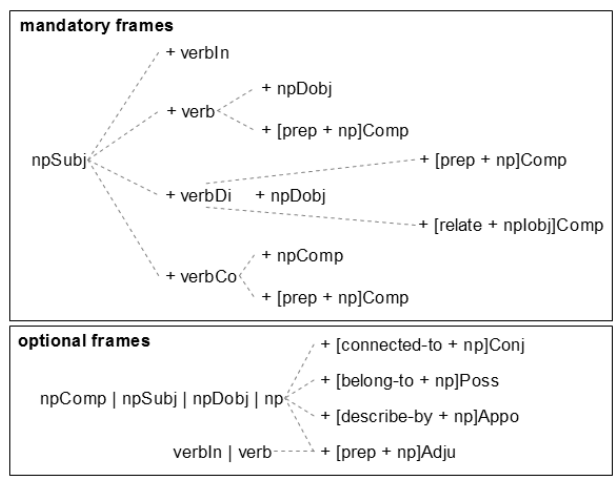}
        \caption{A general set of rules for identifying ternary frames in the form of $<arg1,connector,arg2>$ from parsed texts.}\label{frame}
    \end{center}
\end{figure}

\begin{figure}[thp]
    \begin{center}
\includegraphics[width=4.8in]{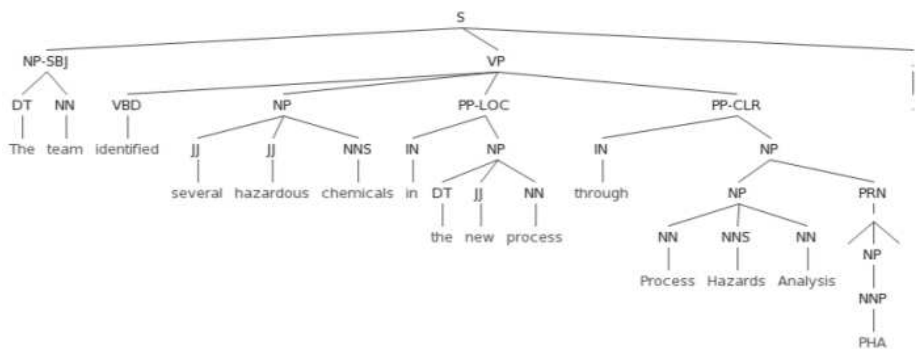}
        \caption{The dependency structure, in the form of  a parse tree for the sentence \emph{``The team identified several hazardous chemicals in the new process through Process Hazards Analysis (PHA)."}.}\label{dependencystructure_eg}
    \end{center}
\end{figure}

\begin{figure}[thp]
    \begin{center}
\includegraphics[width=5.4in]{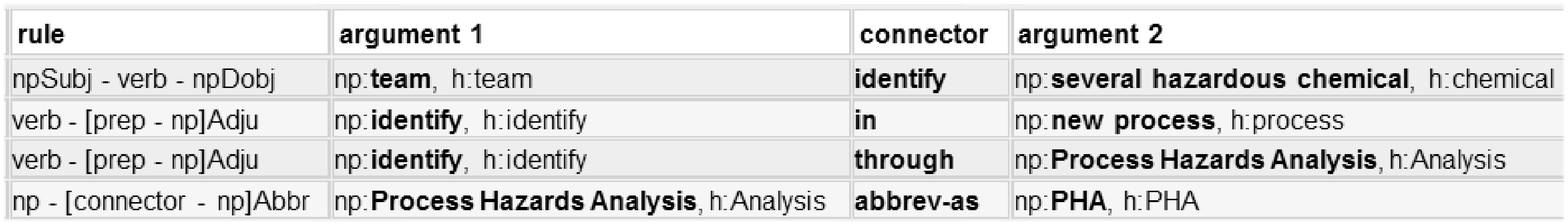}
        \caption{The four ternary frames extracted from the dependency structure shown in Figure \ref{dependencystructure_eg} based on the rules in Figure \ref{frame}.}\label{frame_eg}
    \end{center}
\end{figure}

Next, domain texts, which can be a selected portion of the domain corpus, are then fed into the text processing phase. The text processing phase is a combination of various natural language processing techniques to extract three-part structures known as \emph{ternary frames}. The modules involved during text processing are \emph{sentence parsing}, \emph{noun phrase chunking} and \emph{frame extraction}. The sentence parsing step unveils part-of-speech tags and dependency structures to enable the identification and grouping of noun phrases during noun phrase chunking. An example of a dependency structure, in the form of a parse tree is shown in Figure \ref{dependencystructure_eg}. Existing noun phrase chunking techniques typically employ dependency structure analysis and simple word association measures based on static text corpora to identify stable sequences of noun phrases. Such techniques are unable to handle head nouns with post modifiers such as prepositional phrase and conjunctions. For the accurate chunking of noun phrases, we incorporated two measures known as \emph{Unithood (UH)} \cite{wong_et_al_2007b} and \emph{Odds of Unithood (OU)} \cite{wong_et_al_2008a} for determining the collocation strength of word sequences. For example, the two measures help to decide that the phrase \emph{``Hazard and Operability Study"} should remain as an individual stable unit, while the word \emph{``hazard"} and \emph{``risk"} in \emph{``hazard and risk"} should not be combined. The \emph{OU} measure is the probabilistic reformulation of the ad-hoc combination of unithood evidence by \emph{UH}. The frame extraction step then extracts ternary frames in the form of $<arg1,connector,arg2>$. A general set of linguistic rules are used to determine the presence of such frames from the parsed texts. These rules are summarised in Figure \ref{frame}. Using these rules and the example dependency structure in Figure \ref{dependencystructure_eg}, we can extract several ternary frames as shown in Figure \ref{frame_eg}.

Thirdly, noun phrases appearing as arguments in the ternary frames are gathered to form a list of term candidates for further processing. The set of term candidates \emph{\{``team",``several hazardous chemical",``new process",``Process Hazards Analysis"\}} can be obtained from the example in Figure \ref{frame_eg}. The purpose of the term recognition phase is to identify terms from the list of candidates which are relevant to and representative of the domain of interest. The subjective nature of term relatedness makes termhood determination a challenging issue to address. Several measures for determining termhood have been developed in the past with limited accuracy. In this regard, we have developed two measures, namely, \emph{Termhood (TH)} \cite{wong_et_al_2007d,wong_et_al_2008c} and \emph{Odds of Termhood (OT)} \cite{wong_et_al_2007e} for determining the degree of relatedness of terms to a specific domain. These two measures address various issues which were often neglected such as the importance of modifiers in determining the semantics of a term, the difference between the notion of prevalence and tendency, and the role of contextual information in determining termhood. These two measures make use of the distributional behaviour of term candidates within the target domain corpus and also across other domains (i.e. contrastive corpus) as statistical evidence to quantify various important linguistic evidence. Our two termhood measures have been shown to perform with high accuracy in comparison to two other existing measures \cite{wong_et_al_2008c}. 

The final phase in our ontology construction system is relation extraction. In this phase, we discover unnamed relations between terms through term clustering using an algorithm known as \emph{Tree-Traversing Ant (TTA)} \cite{wong_et_al_2007c,wong_et_al_2008b}. $TTA$ is a hybrid technique inspired by ant-based method \cite{handl_et_al_2003} and conventional hierarchical clustering. $TTA$ is capable of further distinguishing hidden structures within clusters and is tolerant to differing cluster size. The ability to identify and isolate outliers, and to produce consistent results makes $TTA$ a reliable term clustering technique. Unlike conventional systems which require the computationally intensive task of feature extraction and selection, our system employs two featureless measures, namely, \emph{n$^\circ$ of Wikipedia (noW)} \cite{wong_et_al_2006c} and \emph{Normalised Google Distance (NGD)} \cite{cilibrasi_vitanyi_2007} for similarity and distance measurement. Using the relations extracted from this phase, we can then organise the flat list of domain terms discovered from term recognition into a graph structure to produce a lightweight domain ontology. This graph structure can be converted into various format using languages such as OWL. This ontology can be edited and maintained using ontology editors such as $Prot\acute{e}g\acute{e}$ \cite{musen_et_al_1993}.

\section{Evaluations and Discussions}\label{experiment}

For the evaluation, we employ a dataset containing a domain corpus describing \emph{``risk management"} and a contrastive corpus. We constructed a set of about $7,600$ documents extracted from \emph{ScienceDirect} using $34$ keywords provided by experts. These documents together with the electronic version of a risk management textbook contribute to the domain corpus. The contrastive corpus is comprised of about $28,000$ news articles crawled from \emph{Reuters}, \emph{CNet} and \emph{Discovery}, and five off-the-shelf corpora namely \emph{GENIA} \cite{kim_et_al_2003}, \emph{BioCreative}  \cite{hirschman_et_al_2005}, \emph{BioMedCentral}  \cite{cockerill_2003}, \emph{Reuters-21578} \cite{rose_et_al_2002} and \emph{British National Corpus} \cite{burnard_2007}. Table 1 summarises the dataset used in our evaluation. We use the risk management textbook, which is part of the domain corpus, as input to our ontology construction system due to the authority and reliability of its content. We feed the textbook into our text processing phase to obtain ternary frames in the form of $<arg1,connector,arg2>$. Over $22,000$ ternary frames were extracted from the textbook. For practical reasons, we randomly selected $4,000$ frames for further processing. A list of term candidates is then constructed from the $4,000$ ternary frames by selecting distinct noun phrases from $arg1$ and $arg2$. The resulting set $TC$ contains $2,841$ term candidates. In the following two subsections, we will briefly discuss the process of term recognition and term clustering, and the evaluation results associated to each phase.

\begin{figure}[htp]
		\textbf{Table 1. The two types of corpora used in our evaluation, and their content.}
    \begin{center}
\includegraphics[width=5.4in]{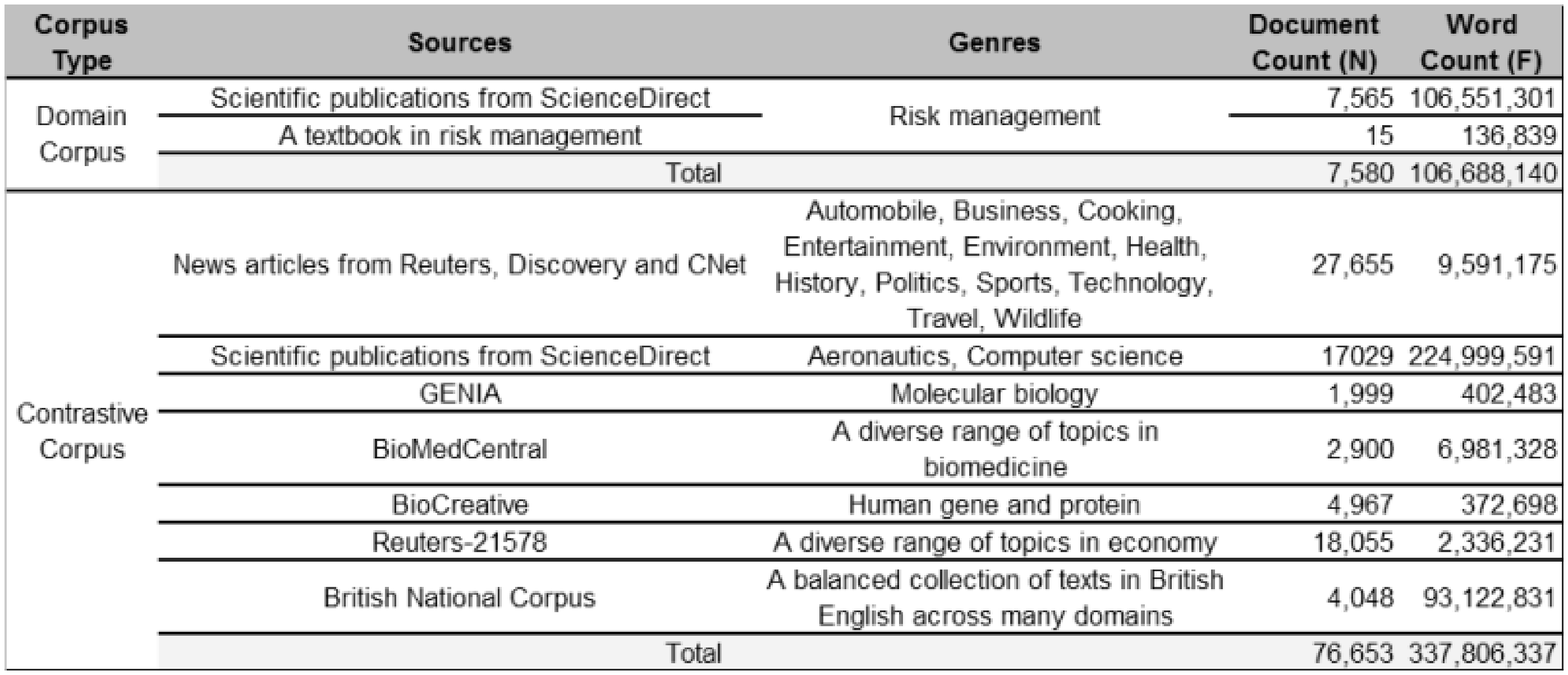}
    \end{center}
\end{figure}

\subsection{Performance of Term Recognition}
In the second phase of ontology construction, we perform term recognition using four termhood measures, namely, $OT$ \cite{wong_et_al_2007e}, $TH$ \cite{wong_et_al_2007d}, $CW$ \cite{basili_et_al_2001} and $NCV$ \cite{frantzi_ananiadou_1997} on the set of $2,841$ term candidates. In this phase, the term candidates are systematically assessed and assigned weights to reflect their relevance to the domain represented by the domain corpus. The term candidates are then ranked according to the weights assigned to them. We evaluated the performance of term recognition from two perspectives. We conducted a qualitative evaluation by analysing the frequency distribution of the terms to determine if terms are properly weighted according to their distributional behaviour across different domains. The second evaluation examines the performance of term recognition quantitatively with the help of domain experts.

\begin{figure}[t]
  \centering
	\subfigure{\label{} \includegraphics[width=2.8in]{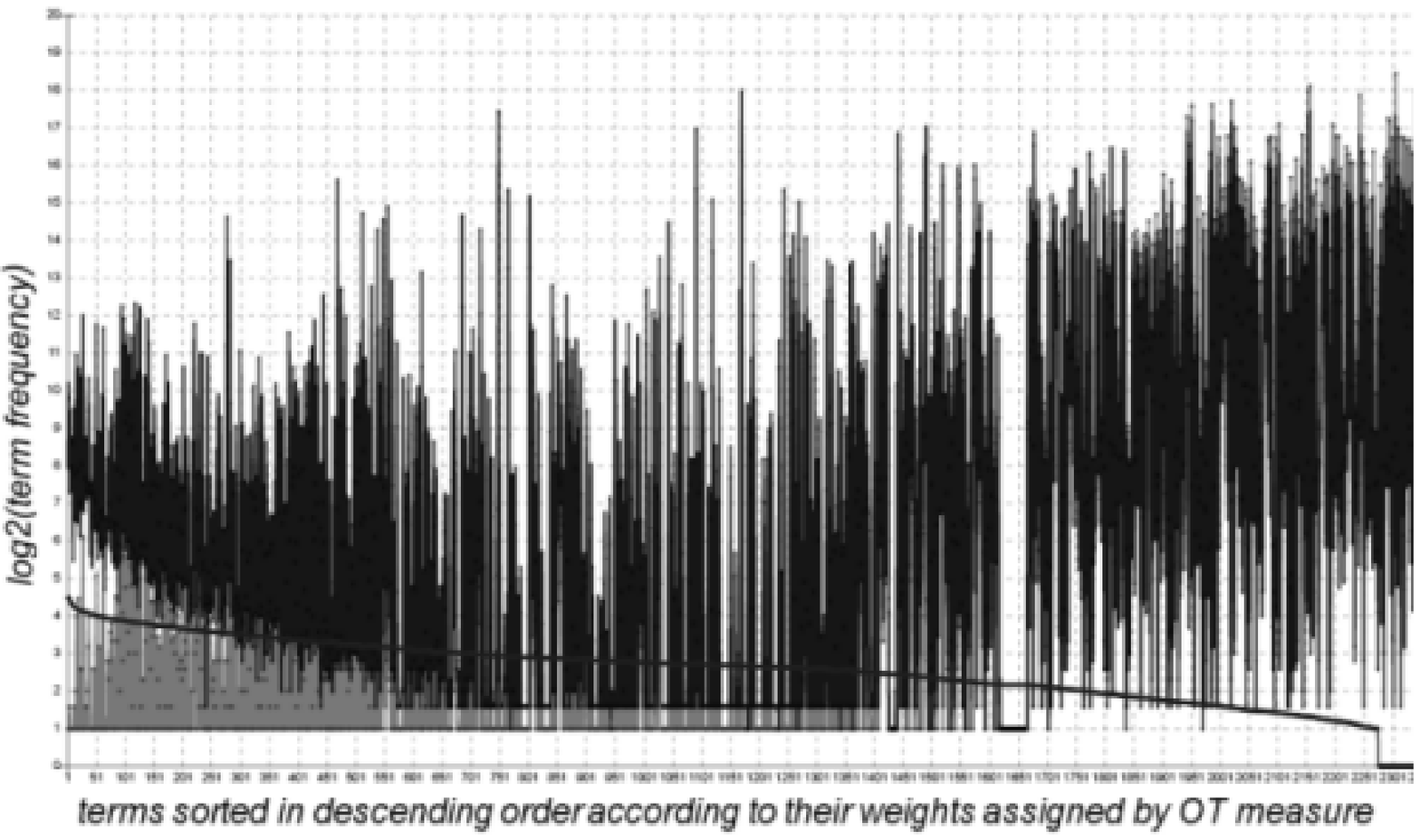}}
	\subfigure{\label{} \includegraphics[width=2.8in]{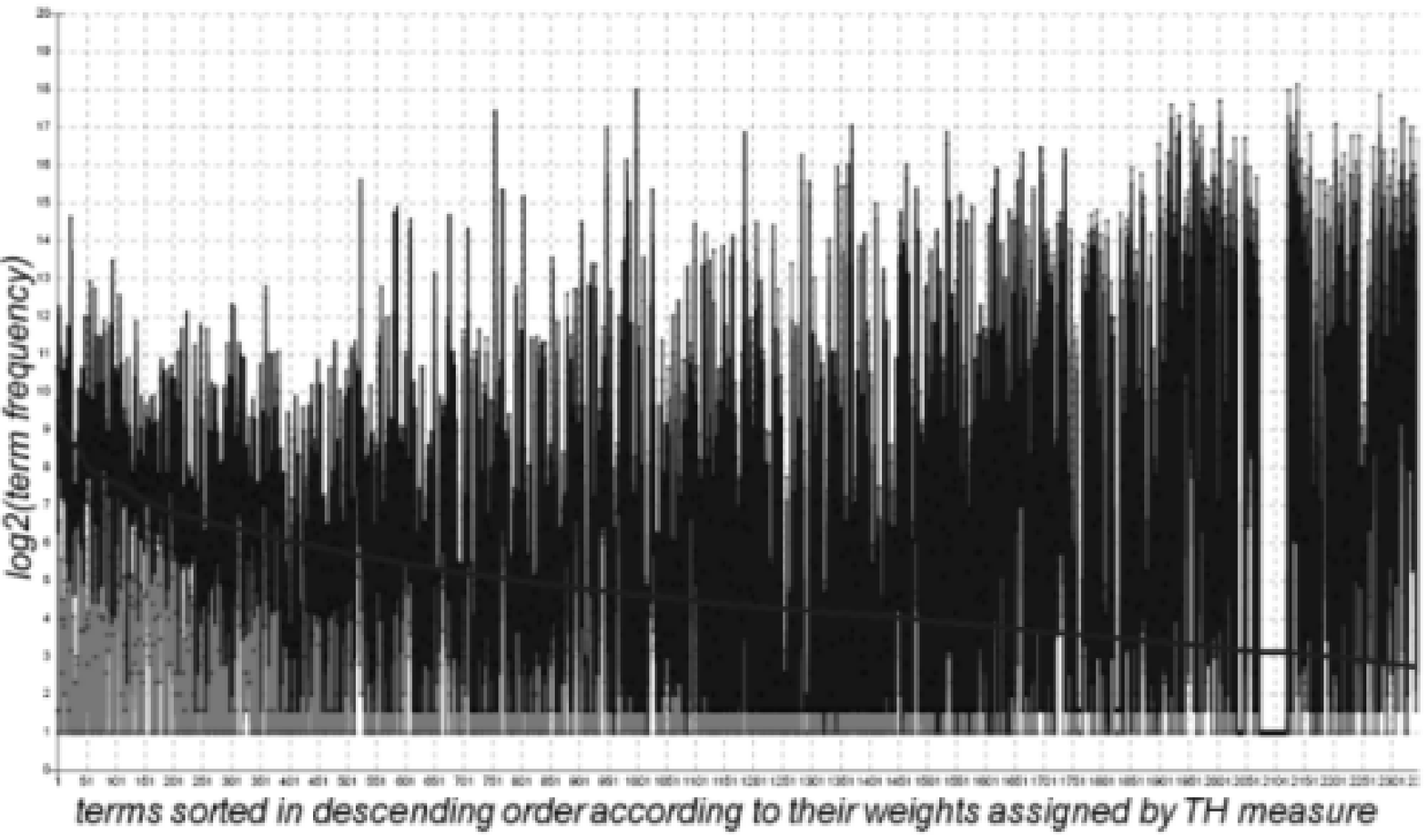}}  			
  \subfigure{\label{} \includegraphics[width=2.8in]{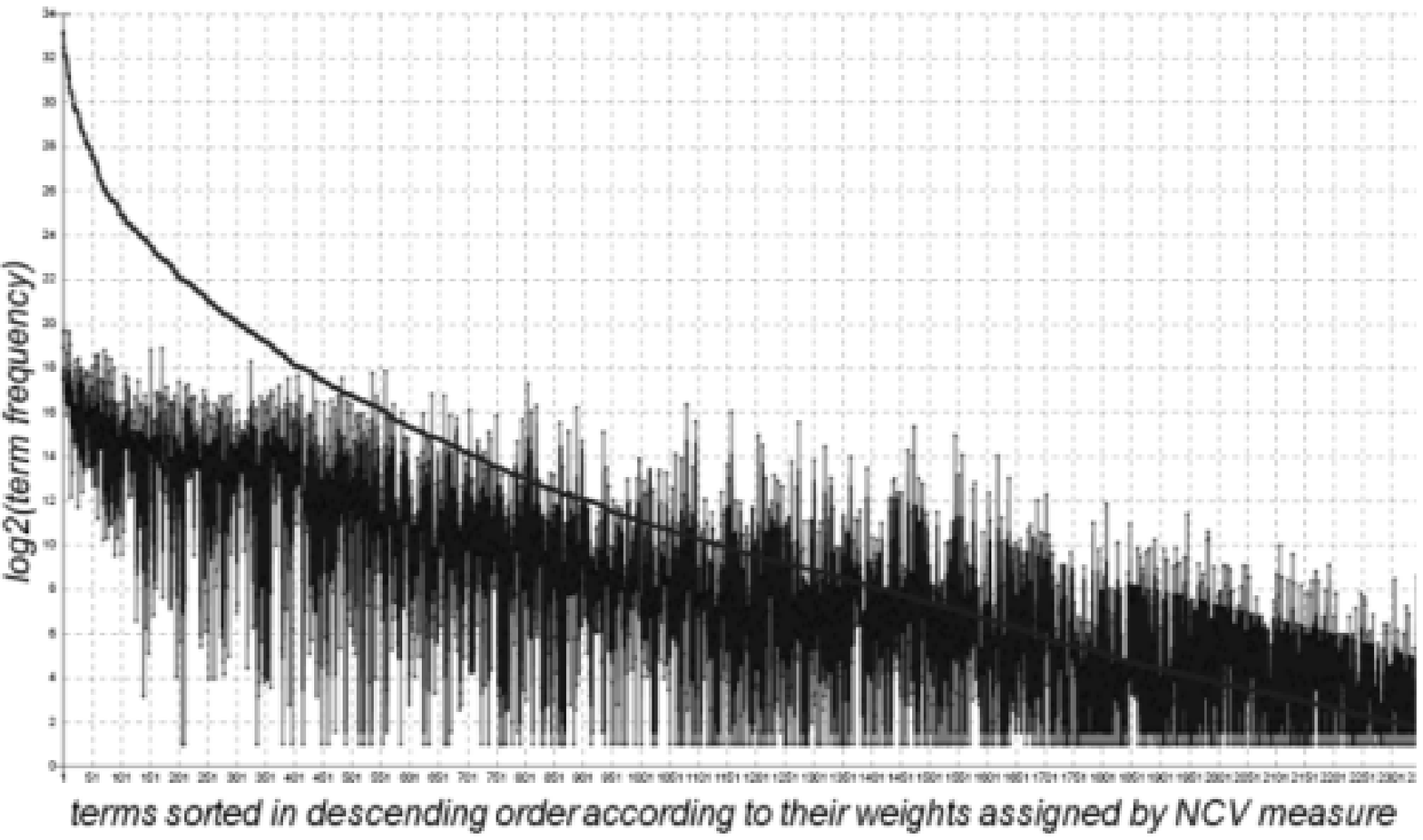}}
  \subfigure{\label{} \includegraphics[width=2.8in]{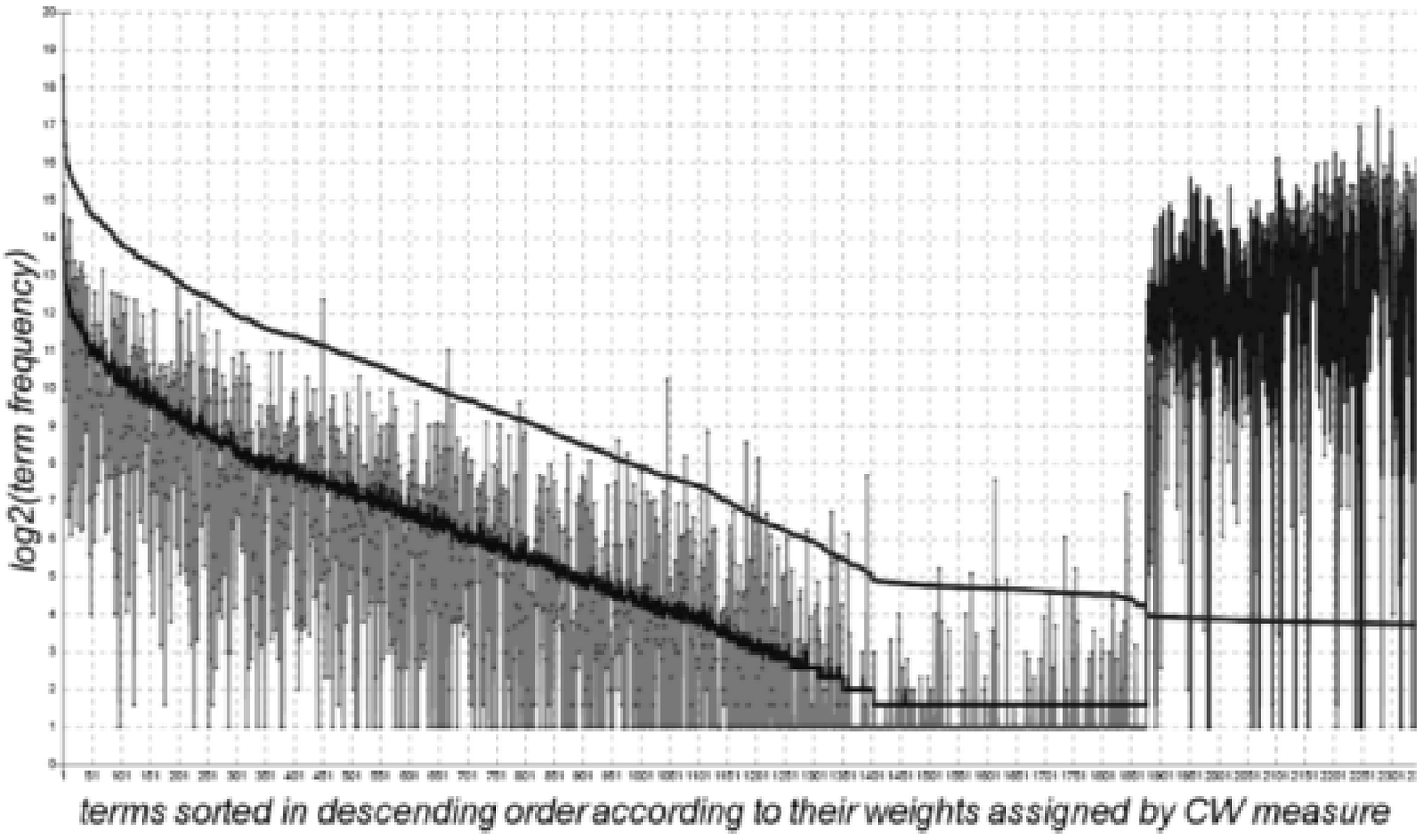}} 	
   \caption{Distributions of the $2,841$ terms extracted from the textbook, which is part of the domain corpus, sorted according to the corresponding scores provided by the four measures. The single dark smooth line stretching from the left (highest value) to the right (lowest value) of the graph is the scores assigned by the respective measures. As for the two oscillating lines, the dark line is the domain frequencies $f_d$ while the light one is the contrastive frequencies $f_{\bar{d}}$.}
  \label{evaluationPartOne}
\end{figure}

In the first part of the evaluation, we analyse the frequency distributions of the ranked term candidates generated by the four measures. Generally, terms which occur more frequently in the domain corpus than in the contrastive corpus should be assigned higher weights. Other factors such as the domain relatedness of heads and context should also be taken into consideration. Figure \ref{evaluationPartOne} show the frequency distributions of the term candidates ranked in descending order according to the weights assigned by the respective measures. In this evaluation, a measure is considered as capable of identifying highly related terms if the corresponding graph shows high degree of polarisation of the oscillating lines. The dark oscillating lines represent the domain frequencies $f_d$ while the grey oscillating lines are contrastive frequencies $f_{\bar{d}}$. Terms which are assigned higher weights are considered as more relevant and are located to the left of the x-axis. Ideally, terms along the start of the x-axis should have very high domain frequency $f_d$ (i.e. located higher on the y-axis) and relatively lower contrastive frequency $f_{\bar{d}}$ (i.e. lower on the y-axis). One can notice the interesting trends from the graphs by $CW$ and $NCV$ in Figure \ref{evaluationPartOne}. The first half of the graph by $CW$, prior to the sudden surge of frequency, consists of only complex terms (i.e. multi-word terms). The relatively lower word count of complex terms as compared to simple terms explains for such disparity in the frequency distribution produced by $CW$. This is attributed to the biased treatment given to complex terms evident in the formulation of the $CW$ measure. However, priority is also given to complex terms by $TH$ but as one can see from the distribution of candidates by $TH$, such undesirable trend does not occur. One of the explanations is the heavy reliance of frequency by $CW$ while $TH$ attempts to diversify the evidences in the computation of weights. While frequency may be a reliable source of evidence, the use of it alone is definitely inadequate \cite{cabre-castellvi_et_al_2001}. As for $NCV$, Figure \ref{evaluationPartOne} reveals that scores are assigned to candidates by $NCV$ based solely on the domain frequency. In other words, the measure $NCV$ lacks the required contrastive analysis. As we have pointed out, terms can be ambiguous and we must not ignore the cross-domain distributional behavior of terms. In addition, upon inspecting the actual list of ranked candidates, we noticed that higher scores are assigned to candidates which are accompanied by more context words. Another positive trait that $TH$ exhibits is its ability to assign higher scores to terms that occur relatively more frequent in the domain corpus than in the contrastive corpus. This is evident through the gap between $f_d$ (dark oscillating line) and $f_{\bar{d}}$ (light oscillating line), especially at the beginning of the x-axis. One can notice that candidates along the end of the x-axis are those with $f_{\bar{d}} > f_d$. The same can be said about our new measure $OT$. However, the discriminating power of $OT$ is apparently better since the gap between $f_d$ and $f_{\bar{d}}$ is larger and lasted longer. 

\begin{figure}[htp]
		\textbf{Table 2. An example of a contingency table. Note that $|TC|=TP+FP+FN+TN$ where $|TC|$ is the total number of term candidates in the input set.}
    \begin{center}
\includegraphics[width=3.4in]{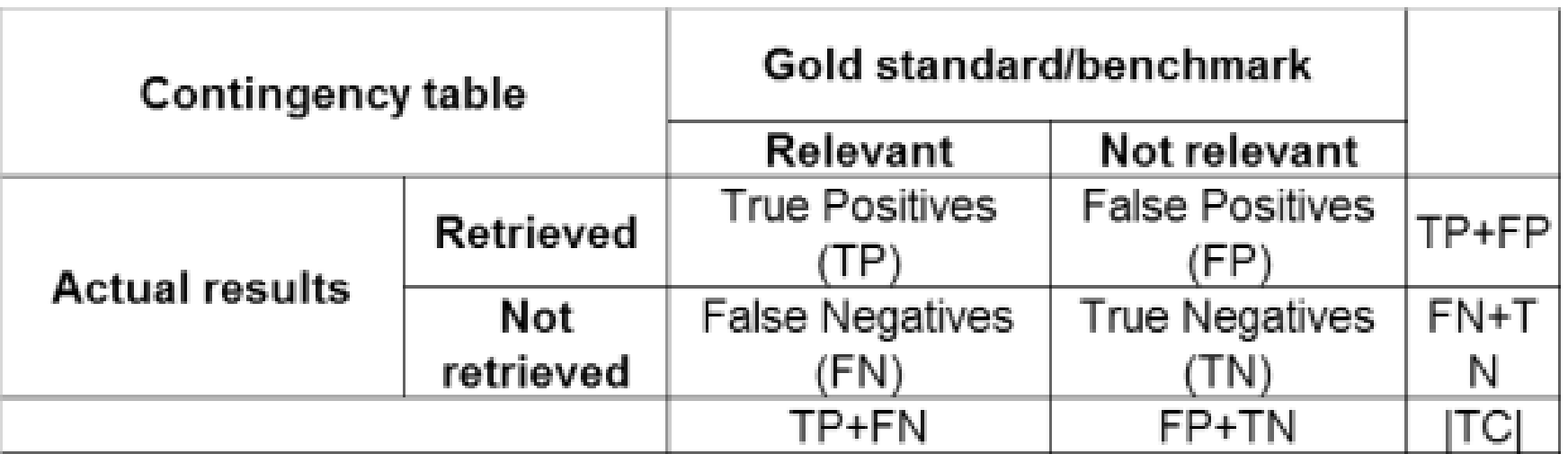}
    \end{center}
\end{figure}

In addition to qualitative assessment, we conducted a quantitative evaluation on the set of terms produced during term recognition with the help of domain experts. There are several performance measures common to the field of Information Retrieval and Information Extraction such as \emph{precision}, \emph{recall}, \emph{F-measure}, and \emph{accuracy}. These measures are computed by constructing a contingency table as shown in Table 2:
\begin{align}
precision 	& = \frac{TP}{TP+FP}\nonumber\\
recall 			& = \frac{TP}{TP+FN}\nonumber\\
F_{1} 	& = \frac{2 \times precision \times recall}{precision + recall}\nonumber\\
accuracy 		& = \frac{TP+TN}{TP+FP+FN+TN}\nonumber
\end{align}
where $TP$, $TN$, $FP$ and $FN$ are values from the four cells of the contingency table shown in Table 2. Considering the manpower constraint in manual assessment, we limit our evaluation to only the top $n=300$ ranked term candidates from each list produced by each measure. In this case, the \emph{``retrieved"} portion of the \emph{``actual results"} in the contingency table refers to the top $300$ terms from each measure. The \emph{``not retrieved"} portion in the table cannot be defined since there is no standard set of terms available through benchmarks or gold standards for our domain of interest. Due to the absence of the two pieces of information $TN$ and $FN$, the precision measure is the only applicable performance indicator of term recognition in this paper. The precision at $n$ is the fraction of the top $n$ term candidates that are considered as relevant to the domain. We seek the help of experts in the domain of \emph{``risk management"} to decide if the top $300$ terms by each measure are actually relevant to the domain. The domain experts performed a binary classification by deciding if a term is relevant or not relevant to our domain of interest. We organised the results in Table 3. The $TP$ row shows the number of terms ranked within the top $300$ which are actually relevant to the domain of \emph{``risk management"}. $FP$, on the other hand, contains the number of non-relevant terms. As shown in Table 3, term recognition using the two measures presented as part of our system (e.g. $OT$ and $TH$) offer far more precise results compared to existing measures. Out of the top $300$ terms ranked by $OT$ and $TH$, about $60-70\%$ of them are considered as relevant to the domain of \emph{``risk management"} as compared to only $10-30\%$ precision by other measures. The actual list of term candidates and the expert's evaluation is available on our research site\footnote{http://explorer.csse.uwa.edu.au/research/\_sandbox\_evaluation.pl}. Despite the mediocre performance figure delivered by our two measures in this evaluation, $OT$ and $TH$ performed extremely well in view of the following reasons:
\begin{itemize}
\item Issues such as coverage and sparsity of the domain corpus have tremendous impact on the performance of term recognition. The domain corpus used in this evaluation was constructed automatically from various sources without any attempt to assess its adequacy in these aspects.
\item The term candidates in this evaluation were automatically extracted from real-world texts without human intervention. The text processing phase and specifically, the extraction of term candidates have errors of their own (e.g. incorrect noun phrase chunking). Such errors will inevitably propagate to the next phase of term recognition. 
\item In manual assessment, evaluation results vary depending on the human experts. Our domain experts imposed stringent requirements during their inspections of the ranked terms. The terms are only classified as relevant if the experts are absolutely certain of the terms' significance to the domain of \emph{``risk management"}.
\end{itemize}

\begin{figure}[htp]
		\textbf{Table 3. The performance of term recognition for all four measures in this evaluation. The $TP$ and $FP$ rows contain the number of relevant terms and the number of terms considered as not relevant by the domain experts, respectively.}
    \begin{center}
\includegraphics[width=3.4in]{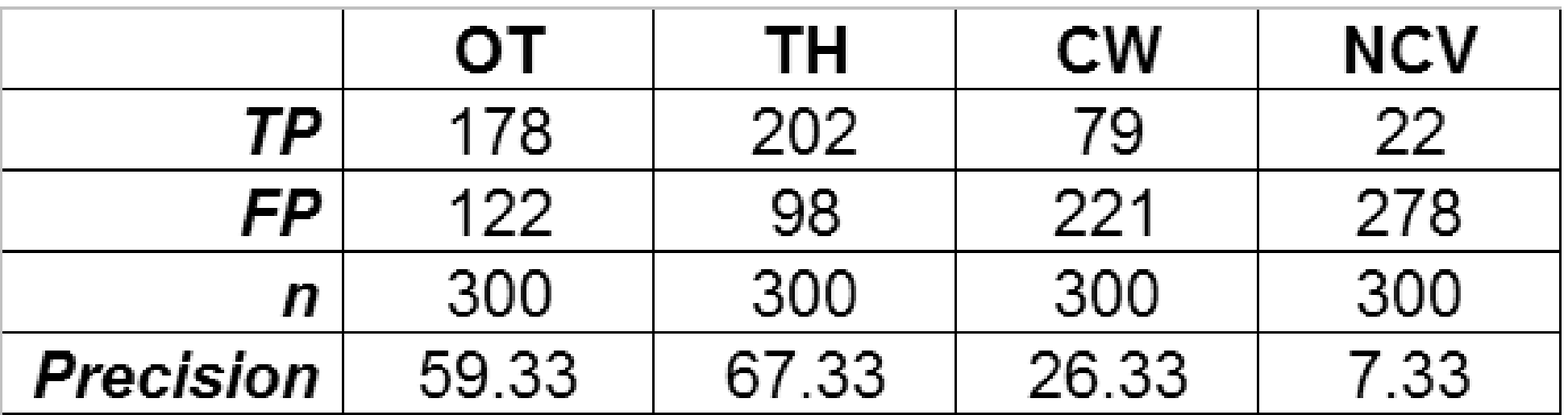}
    \end{center}
\end{figure}

\subsection{Performance of Relation Discovery}
In the second part of the evaluation, we performed relation discovery to identify unnamed associations between the domain terms produced during term recognition. We employed two metrics in ontology learning for evaluation. The first is known as \emph{Lexical Overlap ($LO$)} for evaluating the intersection between the set of discovered concepts ($C_d$) and the benchmark concepts ($C_m$) \cite{maedche_staab_2002}. $LO$ is defined as:

\begin{eqnarray}
\label{}
LO = \frac{|C_{d} \cap C_{m}|}{|C_{m}|}
\end{eqnarray}
where $|C|$ is the number of concepts in set $C$. The second metric known as \emph{Ontological Loss (OL)} is used to identify the number of benchmark concepts that were not discovered during term clusterings. $OL$ is defined as \cite{sabou_et_al_2005}:

\begin{eqnarray}
\label{}
OL = \frac{|C_{m} - C_{d}|}{|C_{m}|}
\end{eqnarray}

\begin{figure}[t]
    \begin{center}
\includegraphics[width=5.4in]{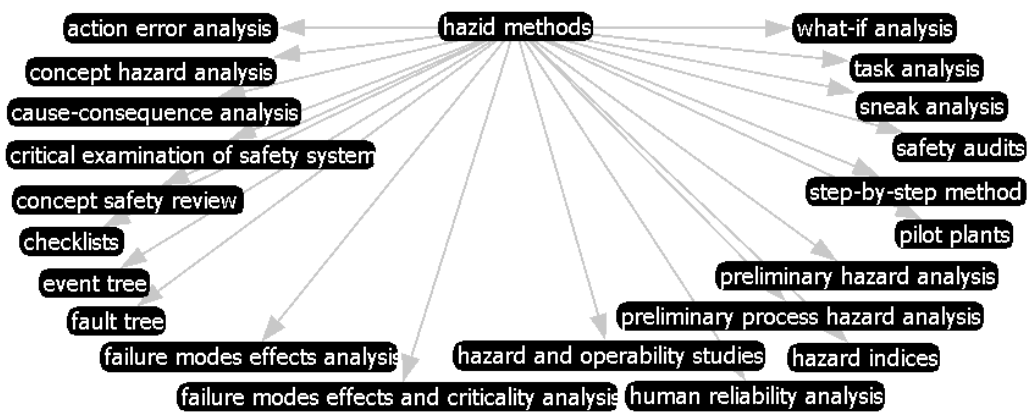}
        \caption{The $21$ upper-level concepts in the hand-crafted domain ontology for hazard identification methods in risk management \cite{gilmour_2004}. These concepts are used as benchmark concepts for evaluation.}\label{ontology_benchmark_toplevel}
    \end{center}
\end{figure}
\begin{figure}[htp]
    \begin{center}
\includegraphics[width=6.2in]{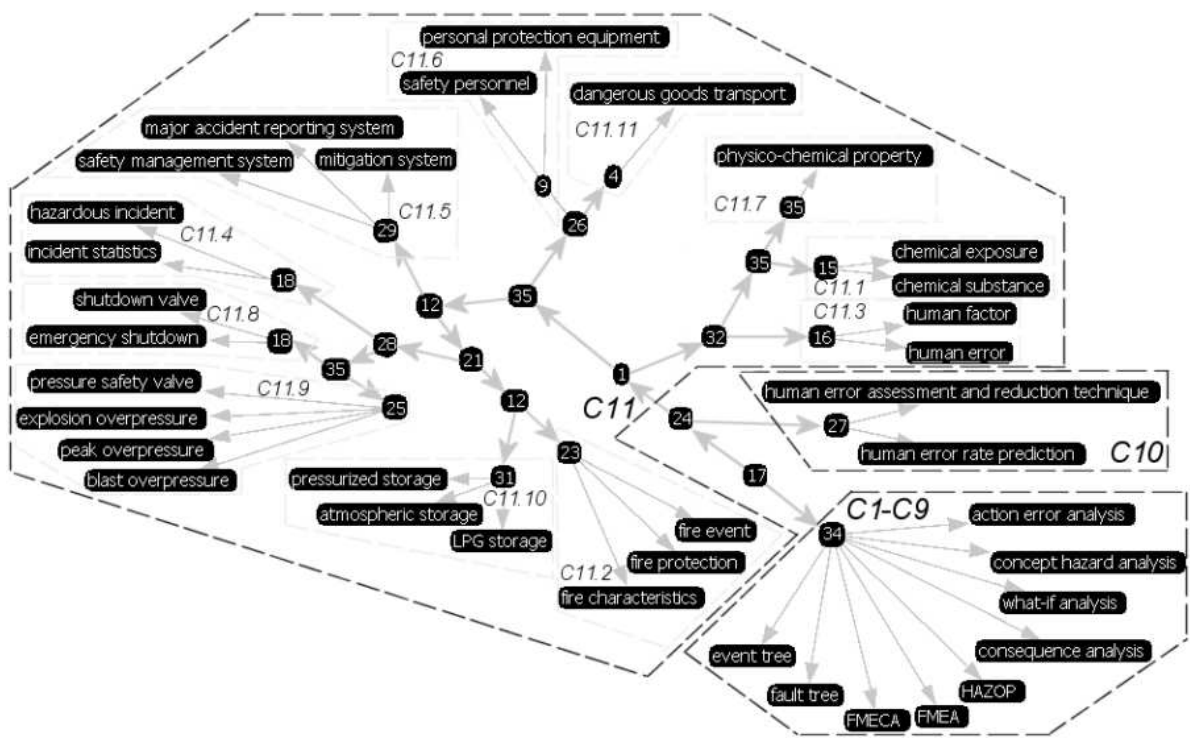}
        \caption{The clustering result produced by our system. The flat list of terms produced by term recognition is organised into a graph structure to form a lightweight domain ontology. The main concepts discovered by the system are labelled as $C1$ to $C11$. These discovered concepts correspond to $11$ out of the $21$ benchmark concepts. Concepts $C1$ to $C9$ are exact matches of the benchmark concepts while $C10$ and $C11$ are concepts discovered through generalising related terms. For instance, concept $C10$ is an abstract representation of \emph{``human reliability analysis"} obtained through generalising related techniques, namely, \emph{technique for ``human error rate prediction" (THERP)} and \emph{``human error assessment and reduction technique" (HEART)}. The concept representing \emph{``checklists"} $C11$ is produced after two levels of generalisation. Terms related to the individual elements of \emph{``checklists"} were first clustered to form the corresponding elements, which in turn were used to discover the main \emph{``checklists"} concept. There were $11$ \emph{``checklists"} sub-concepts (e.g. $C11.1$ to $C11.11$) generated. Sub-concepts $C11.1$ to $C11.11$ represent \emph{``chemical reactors"}, \emph{``fire protection"}, \emph{``human factors \& human errors"}, \emph{``incident investigation"}, \emph{``management system"}, \emph{``personal safety"}, \emph{``physico-chemical property"}, \emph{``plant start-up \& shutdown"}, \emph{``pressure system design"}, \emph{``storage"} and \emph{``transport"}, respectively.}\label{ontology_result}
    \end{center}
\end{figure}
To ensure that high-quality domain ontology is produced, the top $n=36$ terms ranked by $TH$ which has $100\%$ precision were automatically selected for term clustering. We employ a manually-crafted domain ontology for hazard identification methods in risk management \cite{gilmour_2004} as a benchmark for comparison. The benchmark ontology has a set of $21$ concepts as shown in Figure \ref{ontology_benchmark_toplevel}. However, we selected only $16$ concepts from the benchmark ontology for comparison or in other words, $|C_m|=16$. The $5$ concepts excluded were \emph{``concept safety review"}, \emph{``critical examination of safety system"}, \emph{``preliminary process hazard analysis"}, \emph{``pilot plants"} and \emph{``step-by-step method"}. The terms which represent these concepts, and other related terms which can be generalised to form these concepts were not present in the textbook. Such constraints on the part of the textbook can result in the misintepretation of the actual capability of the system proposed in this paper.

The lightweight ontology created using our term clustering technique is shown in Figure \ref{ontology_result}. Our clustering technique managed to discover $11$ out of the $16$ benchmark concepts. For this experiment alone, our system achieved a lexical overlap of $11/16=69\%$. The system experiences a $5/16=31\%$ ontological loss due to the non-discovery of $5$ benchmark concepts. The non-discovered concepts were \emph{``hazard indices"}, \emph{``preliminary hazard analysis"}, \emph{``safety audits"}, \emph{``sneak analysis"} and \emph{``task analysis"}. Upon detail examination, we identified the causes behind the system's inability to discover these $5$ concepts:

\begin{itemize}
\item \emph{``preliminary hazard analysis"} was mentioned only once as plain text (i.e. not part of a figure or table) in the textbook, and was not part of the randomly selected $4,000$ frames for term recognition.
\item \emph{``safety audits"} does not occur in the textbook. Possible related terms such as \emph{``process safety audit"} and \emph{``operational safety audit"} that could assist in discovering a generalised concept have only one to two occurrences in the textbook, and were excluded from the $4,000$ frames.
\item \emph{``sneak analysis"} and \emph{``task analysis"} have low occurrences in the textbook, and were not included for term recognition.
\item \emph{``hazard indices"} was not extracted as part of any frame due to its absence from the textbook. Possible related terms such as \emph{``chemical exposure index"} and \emph{``instantaneous fractional annual loss"} have less than ten occurrences in the book and were excluded from the $4,000$ frames. Other terms such as \emph{``runaway reaction hazard index"} and \emph{``mortality index"} which could help in discovering a generalised concept do not appear in the textbook. 
Another useful term \emph{``fire and explosion index"}, which was mentioned in the book, was not included for term recognition as a complete term due to an error with noun phrase chunking during text processing. The term was extracted as two separate parts \emph{``fire"} and \emph{``explosion index"} in the $4,000$ frames.
\end{itemize}
In short, the three general reasons behind the system's inability to discover the desired concepts are 1) inadequate statistical evidence due to low occurrence in text corpora, 2) incidental exclusion during input at each phase, and 3) errors introduced at each phase. Only the third reason is the result of our system faults. The first reason is due to the inadequacy of text corpora and is not related to the system. The second reason is caused by user-imposed constraints during input where useful information is unintentionally filtered out. For example, in this experiment, we restricted the input of term recognition to only $4,000$ ternary frames. In addition, only the top $36$ ranked terms were further processed to construct the lightweight domain ontology. Referring back to our specific causes above, $4$ out of the $5$ undiscovered concepts in this experiment were the results of the first and second reasons, which are unrelated to our system. Notwithstanding these $4$ concepts (i.e. $|C_m|-4$), our system is in fact capable of performing at an outstanding lexical overlap of $11/(|C_m|-4)=92\%$. In other words, there is only an $8\%$ ontological loss that is actually caused by our system faults.

\section{Conclusions and Future Work}\label{conclude}

Domain ontologies in risk management for chemical processes is becoming increasingly important to support information sharing and reuse among chemical engineers and information systems. In this paper, we presented and evaluated an automatic lightweight ontology construction system based on dedicated text mining techniques. The system is composed of four main phases, namely, text cleaning, text processing, term recognition and relation discovery. The techniques in each phase were designed to employ only dynamic resources such as Wikipedia and Google to ensure applicability to different domains. Our evaluations using real-world texts have shown that the proposed system is capable of automatically constructing high quality lightweight domain ontologies. The performance of lightweight ontology construction using our system can be improved tremendously by 1) diversifying and verifying the sources of term candidates, and 2) increasing the size and improving the quality of text corpora used for term recognition.

The constructed domain ontology is a valuable asset for various applications in risk management. One of such applications is ontology-based document indexing and retrieval. We are planning to employ the domain ontology to perform document indexing on the domain texts. The texts can be indexed using the concepts in the ontology, and related documents can then be easily determined using the relations in the ontology. Such facility offers a conceptual view of the documents in a collection for improving document searchability.

\section*{Acknowledgement}
This research was supported by the Australian Endeavour International Postgraduate Research Scholarship, and the Inter-university Grant from the Department of Chemical Engineering, Curtin University of Technology.

\bibliographystyle{elsart-num-sort}
\bibliography{_reference}

\end{document}